\documentclass[11pt,a4paper]{article}
\usepackage[hyperref]{acl2020}
\usepackage{times}
\usepackage{booktabs}
\usepackage{float}
\usepackage{graphicx}
\usepackage{latexsym}
\usepackage{tabularx}
\usepackage{cleveref}

\restylefloat{table}
\newcolumntype{C}{>{\centering\arraybackslash}X}

\crefformat{section}{\S#2#1#3}
\crefformat{subsection}{\S#2#1#3}
\crefformat{subsubsection}{\S#2#1#3}

\usepackage{microtype}

\aclfinalcopy 

\title{Spelling Correction with Denoising Transformer}
  
\author{Alex Kuznetsov \\
  HubSpot, Inc. \\
  Dublin, Ireland \\
  \texttt{akuznetsov@hubspot.com} \\\And
  Hector Urdiales \\
  HubSpot, Inc. \\
  Dublin, Ireland \\
  \texttt{hector@hubspot.com} \\}

\date{}

\begin{document}
\maketitle
\begin{abstract}
We present a novel method of performing spelling correction on short input strings, such as search queries or individual words. At its core lies a procedure for generating artificial typos which closely follow the error patterns manifested by humans. This procedure is used to train the production spelling correction model based on a transformer architecture. This model is currently served in the {\ifaclfinal{HubSpot}\else{\textit{Company Name}}\fi} product search. We show that our approach to typo generation is superior to the widespread practice of adding noise, which ignores human patterns. We also demonstrate how our approach may be extended to resource-scarce settings and train spelling correction models for Arabic, Greek, Russian, and Setswana languages, without using any labeled data.
\end{abstract}

\section{Introduction}

As search engines in web services rely on user-generated input, they are exposed to a significant degree of noise originating from human error. This leads to 10-15\% of web search queries being misspelled \cite{Dalianis2002,Cucerzan2004}, with percentage of misspellings increasing to up to 20\% for long-tail queries \cite{Broder2009}, and 26\% for academic search engines \cite{Wang2003}. In order to reduce user effort and increase search results recall, spelling correction systems are used. Traditionally such systems use statistical techniques and noisy channel models \cite{Bassil2012,hasan2015spelling,eger2016comparison,gupta2019personalized}. However, in recent years, a number of promising deep learning approaches were developed, spanning applications from e-commerce \cite{zhou2017spelling} to mobile device keyboards \cite{ghosh2017neural}. One downside of deep learning models is their tendency to overcorrect (i.e. correct inputs which should be left as-is) \cite{movin2018spelling,Zhu2019}. This phenomenon results in the network performing corrections in cases when they are not expected: initially correct or, conversely, niche or completely gibberish queries.

The main application of our work is the {\ifaclfinal{HubSpot}\else{\textit{Company Name}}\fi} search, which is used to find contacts, companies, documents, and many other types of items. This means that spelling correction for it has to support inputs in any language, case, containing punctuation, and special characters. Therefore, it is reasonable to treat a search query as a single entity, which can be composed of any set of UTF-8 characters. This frames the task at hand as a query-to-query problem, with a user’s query being an input, and a correction (or the same query in absence of a correction) being the output. Such problem setting naturally leads us to a deep learning implementation of a spelling corrector, in particular using a model with the transformer architecture \cite{vaswani2017attention}.

We stress the importance for the model to produce outputs that are identical to its inputs in a low-confidence setting (unfamiliar or niche query, noisy input, etc). This feature allows us to serve query corrections at a very high precision even on queries containing unique and previously unseen tokens, without the overcorrecting behaviour mentioned above.

Additionally we show that, when combined with the ability to generate an infinite number of realistic (i.e. not simply uniformly random) typos in any language which can be mapped to the QWERTY keyboard, this approach allows to train robust spelling corrections systems for any setting, regardless of the volume of labeled data available. We illustrate this by training spelling correctors for Arabic, Greek, Russian, and Setswana languages, without using any misspelling examples in these languages.

\section{Related Work}

Previous research suggested framing spelling correction as a string-to-string translation or a sequence-to-sequence problem \cite{hasan2015spelling,eger2016comparison,zhou2017spelling,movin2018spelling,wang2019confusionset,zhang2019investigation}. In recent years deep learning approaches to spelling correction were actively explored \cite{sun2015convolutional,zhou2017spelling,ghosh2017neural,etoori2018automatic,li-etal-2018-hybrid,movin2018spelling}, including applications \cite{zhang2019investigation,grundkiewicz2019neural} of a transformer architecture \cite{vaswani2017attention}. Several works highlight the overcorrection problem, when the model underestimates the self-transformation probability \cite{sun2010learning,Zhu2019}. Lack of sufficient training data (noisy to correct mappings) is another important problem \cite{ghosh2017neural}.

Introduction of artificial noise was previously explored in order to overcome low volume of training data \cite{hasan2015spelling,etoori2018automatic,wang2019denoising,choe2019neural,grundkiewicz2019neural}. However, to the best of our knowledge, our approach is the first to generate character-level noise using statistics derived automatically and purely from human error patterns and which combine typo type frequencies, typo probability given position in a string, and character confusion sets. We avoid using Gaussian distributions and human-engineered heuristics and, instead, derive all statistics exclusively from search logs data. \newcite{etoori2018automatic} derive human error patterns but no detail is provided about deriving patterns beyond error types. Character confusion sets were used before, predominantly in a Chinese language setting \cite{liu2013hybrid,chen2013study,wang2018hybrid,wang2019confusionset}. Word level confusion sets were studied as well, focusing on grammatical error correction \cite{wang2019denoising,choe2019neural,grundkiewicz2019neural}, preposition usage \cite{rozovskaya2010generating}, and dyslexic errors \cite{pedler2010large}.

Search engine queries may serve as a useful resource for development of spelling correction models  \cite{Cucerzan2004,gao2010large}. It is common to use search engine logs in order to collect user-issued query rewrites as labels for training spelling correction systems \cite{radlinski2005query,zhang2006discriminative,hasan2015spelling,Zhu2019}. Some researchers \cite{gao2010large,sun2010learning,movin2018spelling} find clicks on query suggestions to be another reliable source of labels.

As an alternative to custom deep learning spelling correction models, statistical open-source models may be used, such as SymSpell\footnote{\url{https://github.com/wolfgarbe/SymSpell}}. We evaluated symspellpy\footnote{\url{https://github.com/mammothb/symspellpy}} and spelling corrector by Peter Norvig\footnote{\url{https://norvig.com/spell-correct.html}} as examples of such models. When tested on a dataset of {\ifaclfinal{HubSpot}\else{\textit{Company Name}}\fi} search logs, we found the following disadvantages, with some of them highlighted by authors of these models: 1) on noisy domains like search logs such models overcorrect at the very high rate (i.e. gibberish or incomplete search queries tend to get corrected towards known vocabulary words when it is not desired), 2) absence of a confidence score for model outputs (it is possible to use proxies such as edit distance and token popularity, however such proxies are too coarse), 3) such models can not be trained or tuned. We, therefore, focus on a deep learning approach to address these shortcomings.

\section{Generating Realistic Typos}
Training deep learning spelling correction models is typically based on a large corpus of \texttt{<typo\_string, correct\_string>} pairs. Such pairs can either be collected \cite{hasan2015spelling,movin2018spelling} or generated artificially \cite{felice2014generating,rei2017artificial}. In our case we are constrained by a dataset of 195K unique \texttt{<typo\_string, correct\_string>} pairs which is insufficient for training a sequence-to-sequence model which is capable of generalising to the diversity of data seen at inference time. As our experiments have shown, a model trained on such a small dataset will suffer from the overcorrection behaviour highlighted in earlier studies \cite{sun2010learning,Zhu2019}.

We address these challenges by reverse engineering the way humans make typos, and using this knowledge to generate as many typos as needed on our unlabeled dataset. This dataset is then used to train a denoising autoencoder which learns to attempt corrections only on misspellings of commonly used words, ignoring unfamiliar queries (which can be typos, gibberish or valid long-tail searches).

There are three main parts to the construction of a training dataset: typo mining (\cref{sec:typo_mining}), typo stats extraction (\cref{sec:typo_stats_extraction}), and typo generation (\cref{sec:typo_generation}).

\subsection{Typo Mining\label{sec:typo_mining}}
We use unique search queries issued in {\ifaclfinal{HubSpot}\else{\textit{Company Name}}\fi} product to look for pairs of similar queries which were fired by the same user close to each other in time (we use a rolling window of 10 subsequent queries). Such pairs often are user-issued query rewrites which may happen due to a spelling mistake. In order to maximise the chance that one query qualifies as a typo of another query, the following set of rules is applied:
\begin{itemize}
\item There is a small edit distance between two queries. We allow for maximum Damerau-Levenshtein edit distance \cite{damerau1964technique} of 1.
\item There is a significant (at least 15X) difference in popularity of two queries (i.e. we assume the correct query is much more popular).
\item The query is considered to be “correct” if all its tokens are either present in the verified vocabulary (list of known names and English words), or belong to 1.5K most popular tokens in search logs.
\item The candidate “typo” query is not composed solely of known tokens (this excludes cases of alternative name spellings).
\item Queries do not contain any forbidden special characters (e.g. \texttt{@}, \texttt{\_}, \texttt{\#}, \texttt{\textbackslash}).
\item The candidate typo query is not a prefix of a correct query (e.g. excluding pairs such as \texttt{<jac, jack>}, \texttt{<jess, jessica>}, \texttt{<mick, mickey>}).
\item Correct query is not a part of a candidate typo query (e.g. excluding pairs such as \texttt{<jimmy, jim>}, \texttt{<alex, lex>}, \texttt{<anastasia, stas>}).
\end{itemize}
Applying these filters on 94M search queries containing 135M tokens (19M unique) produces a collection of 195K \texttt{<typo\_string, correct\_string>} pairs composed of 296K tokens (210K unique).

\subsection{Typo Stats Extraction\label{sec:typo_stats_extraction}}
Using the \texttt{<typo\_string, correct\_string>} pairs from the previous step we extract typo-related statistics. These include: typo type (insertion, substitution, deletion, transposition) frequencies, character confusion matrix (i.e. probability of an accidental swap of any two characters), and distribution of normalised typo positions in a string. This information describes some of the main patterns of how humans make typographical errors, effectively taking into account keyboard and, in part, phonetic proximity.

\textbf{Types of typos.} We follow the commonly used convention of classifying string edits into four categories: insertion, substitution, deletion, and transposition. These categories account for 80-95\% of all typos \cite{Martins2004}. Number of typos in each category for our dataset is presented in Table~\ref{tab:typo_cnts_table}.

\begin{table}[H]
\centering
\begin{tabular}{lll}
    \toprule
    Typo Type&Number of Pairs&\% of Total\\
    \midrule
    Insertion&64060&32.74\\
    Substitution&75922&38.80\\
    Deletion&34580&17.67\\
    Transposition&21103&10.79\\
    \midrule
    Total&195665&100.00\\
    \bottomrule
\end{tabular}
\caption[Number of examples per typo type.]{Volume of examples for each typo type.}
\label{tab:typo_cnts_table}
\end{table}

\begin{figure}[htbp]
\resizebox{\columnwidth}{!}{%
\includegraphics{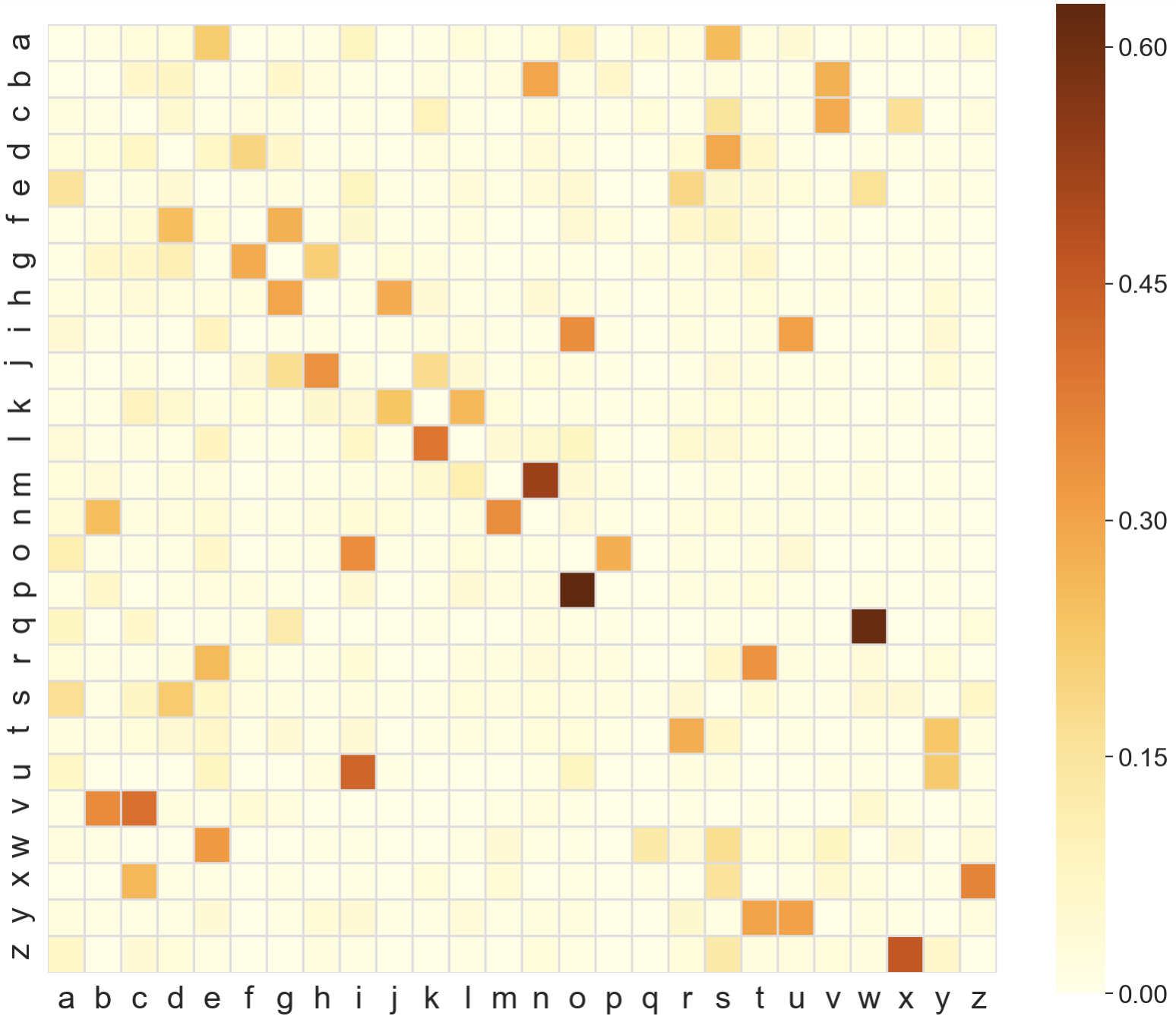}%
}
\caption{Character confusion set. Restricted to lower case English alphabet for visualisation. Character pairs which are close to each other on a keyboard tend to have higher values. Values in each row sum up to 1.}
\label{fig:confusion_set}
\end{figure}

\textbf{Character confusion set.} We find that \texttt{<typo\_string, correct\_string>} pairs which belong to a \textit{Substitution} category are a reliable source of information behind character-level errors. In particular, we are able to derive, for each character, a probability distribution (over all other characters) of making an erroneous substitution. These distributions highlight that keyboard proximity and phonetics are significant drivers of typing errors. We illustrate these findings in Figure \ref{fig:confusion_set} with a character confusion set for lower case English alphabet with all other characters excluded for visualisation purposes. Full confusion set contains 75 characters misspelled for 208 characters.

\textbf{Position of a typo in a string.}
The probability of making a spelling mistake is a function of its normalised position within a string. 

We start by finding the first position at which the correct query and the typo query differ, and divide this position by the length of the correct query.

Normalising typo positions for all \texttt{<typo\_string, correct\_string>} pairs allows us to compute a probability of making a typing mistake by each of 100 percentiles (e.g. probability at 66th percentile corresponds to a probability of making a typo in first 2/3 of the string). Based on an input string length we convert probabilities for 100 percentiles to a probability mass function over all character positions in a string. With typo probabilities assigned to each individual character in an input string, it is trivial to iterate over such string and generate typos, following the patterns exhibited by humans.

\begin{figure}[htbp]
\resizebox{\columnwidth}{!}{%
\includegraphics{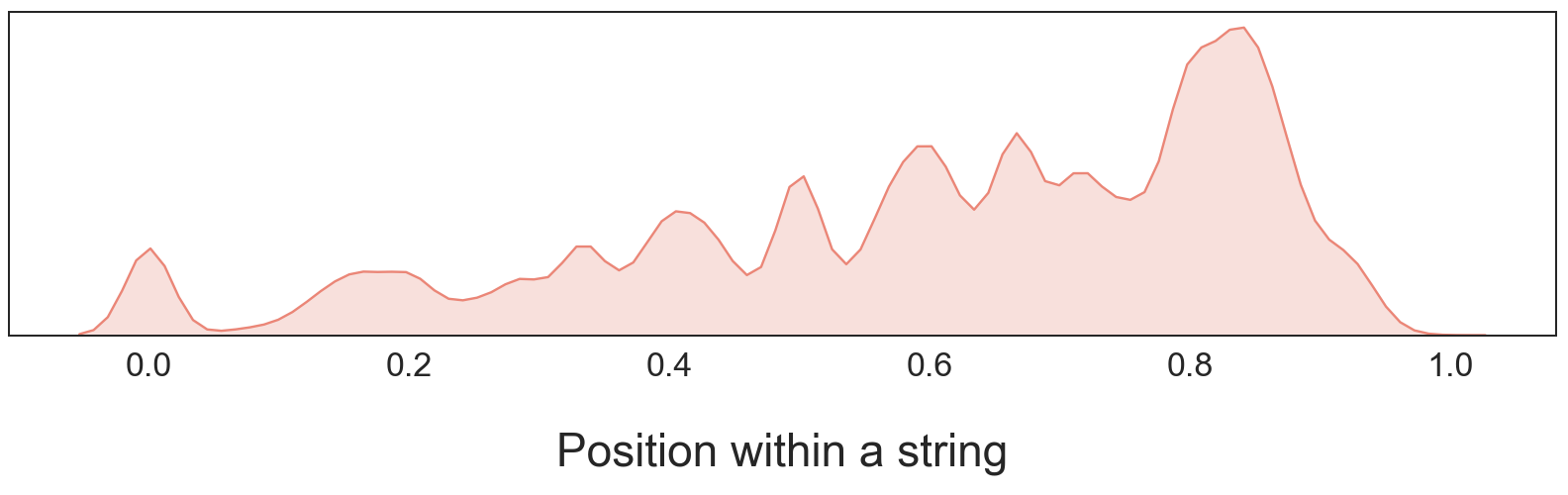}%
}
\caption{Distribution of a normalised typo position within a string.}
\label{fig:string_pos_dist}
\end{figure}

We find that typos tend to happen closer to the end of the string, confirming findings in earlier studies \cite{en_spelling}. The distribution of normalised typo positions is presented in Figure \ref{fig:string_pos_dist}.

\subsection{Typo Generation\label{sec:typo_generation}}
Using the statistics described above, we are able to generate any number of realistic typos that closely mimic human errors. Our algorithm accepts any string as an input and generates a realistic-looking typo, based on string length and statistics described above. We run this algorithm directly on search logs, attempting typo generation for each record. On average we introduce 1 typo per record, but due to the stochastic nature of the typo generation procedure, some records may have 0 or 2 typos.

Using this procedure we generate a dataset of 94M examples, which is significantly larger than the labeled records we could obtain with any other method. A byte pair encoding (BPE) tokenizer \cite{sennrich2015neural} is then fit on this dataset, resulting in a vocabulary of 12K tokens. Similar to \newcite{zhou2017spelling}, we find subword tokenization to be the best way of representing model inputs. The whole dataset is tokenized using the BPE tokenizer, shuffled, and split in 100:1:1 proportions into train, validation, and test sets. This approach allows us to have a training set that is very similar to the data seen in production, thereby minimising possible distribution drifts.

One side effect of constructing \texttt{<typo\_string, correct\_string>} pairs directly from search logs is that noise is introduced over already erroneous queries, like gibberish, niche or partial searches. This may appear detrimental to model performance, however, surprisingly, we observe that in practice noise introduced over erroneous queries does not hurt the quality of model outputs. Instead, given the large size and diversity of the dataset, it forces the model to output sequences identical to the input sequence by default, and to only attempt correction in cases of high certainty (e.g. if a typo is made in a sufficiently popular token). By forcing model outputs to be identical to model inputs in case of gibberish queries, this setup effectively addresses the overcorrection problem \cite{movin2018spelling,Zhu2019}.

Additionally, as some input tokens are compounds (e.g. email addresses, containing first and last names, domain address, etc), this setup forces the model to handle multiple typos in several distinct entities within a single contiguous string.

The ability to train a well performing model directly on the noise generated over unprocessed search logs is surprising, and to the best of our knowledge was not demonstrated before.

\section{Spelling Correction Model \label{sec:spelling_correction_model}}

We train a denoising autoencoder transformer model to recover the original query (without noise) from the query which potentially contains a typo. As model inputs and labels are generated directly from logs data, the distribution of queries is similar between training and inference settings, thereby minimising distribution drifts and biases. Additionally, as queries are seen by the model according to their popularity, the model will naturally learn most frequent terms and will be forced to learn to ignore (and not correct typos on) infrequent, often erroneous and incomplete queries. The production version of our model is a transformer with 4 layers, 2 attention heads, hidden layer size of 256, trained for ~1.6M steps with default learning rate warm-up and decay. This results in a model with 10M trainable parameters. For model implementation we rely on a \texttt{tensor2tensor}\footnote{\url{https://github.com/tensorflow/tensor2tensor}} library \cite{tensor2tensor} and use the hyper-parameter set defined
in the library as \texttt{transformer\_small}).

\section{Experiments}
Below we present two experiments: one comparing approaches to artificial noise generation, and another demonstrating ability to perform transfer to other languages for which no labels are available. 
Maximising model quality was not our goal in these experiments, and we expect that additional tuning of the model, vocabulary generation and training procedures, as well as using beam search score as a confidence threshold will yield significant improvements in quality of spelling correction.
\subsection{Realistic vs Uniformly Generated Typos}
\label{sec:real_vs_uni}
We compare two approaches of generating training data: using realistic typos (\textit{Real}) and using a baseline (\textit{Base}) approach which generates typos in a uniformly random way. For \textit{Real} approach we use typo statistics derived from search logs and for \textit{Base} approach all typo types and string positions are treated as equally probable, and characters for \textit{Insertion} and \textit{Substitution} are chosen uniformly at random from a set of lowercase and uppercase English alphabet letters. For this comparison we train identical denoising transformer models on artificially generated typos for two datasets: {\ifaclfinal{HubSpot}\else{\textit{Company Name}}\fi} search logs (94M original search queries, not tokenized) and a dataset of 3036 Project Gutenberg books (tokenized into 51M tokens, 989K unique) \cite{lahiri:2014:SRW}. For each dataset we generate both uniform and realistic versions of a typo for exactly the same set of input strings. Apart from tokenization for the Gutenberg dataset, no data preprocessing is performed. Models trained on the Gutenberg dataset are evaluated on ground truth datasets of English typos: Wikipedia Common Misspellings\footnote{\url{https://en.wikipedia.org/wiki/Wikipedia:Lists_of_common_misspellings/For_machines}},  and Aspell, Birkbeck, Holbrook datasets\footnote{\url{https://www.dcs.bbk.ac.uk/~ROGER/corpora.html}}. Models trained on {\ifaclfinal{HubSpot}\else{\textit{Company Name}}\fi} search logs are evaluated on a dataset of 195K \texttt{<typo\_string, correct\_string>} pairs described in section \cref{sec:typo_mining}. All models are identical to the one described in section \cref{sec:spelling_correction_model} and are trained for 200K steps (5-6 hours on Nvidia V100 GPU). We report sequence-level accuracy on both \texttt{<typo\_string, correct\_string>} (\textit{Typos}) and \texttt{<correct\_string, correct\_string>} (\textit{Identity}) pairs. Accuracy on \textit{Identity} pairs is equivalent to ${1-FPR}$, where $FPR$ is False Positive Rate\footnote{\url{https://en.wikipedia.org/wiki/False_positive_rate}}. Results of this experiment are presented in Table \ref{tab:realistic_vs_uniform}.

\begin{table}[H]
\centering
\begin{tabular}{{l}*{4}{c}}
    \toprule
    & \multicolumn{2}{c}{Typos} & \multicolumn{2}{c}{Identity}\\
    \cmidrule(lr){2-3}
    \cmidrule(lr){4-5}
    Dataset&Real&Base&Real&Base\\
    \midrule
    Search Typos&\textbf{56.84}&43.70&96.09&96.83\\
    \midrule
    Wikipedia&\textbf{65.92}&63.58&84.90&86.39\\
    Aspell&\textbf{40.30}&37.66&83.78&84.22\\
    Birkbeck&\textbf{33.34}&29.27&85.14&85.40\\
    Holbrook&\textbf{17.92}&17.25&73.92&74.92\\    
    \bottomrule
\end{tabular}
\caption[Comparison of realistic and uniformly random typo generation approaches.]{Comparison of realistic and uniformly random typo generation approaches.}
\label{tab:realistic_vs_uniform}
\end{table}

Experiment results suggest that there is a considerable benefit in generating typos in a realistic manner, which is especially evident in the case of our in-house search typos dataset, from which human error patterns were derived. The fact that error patterns derived from search typos may be successfully transferred to other domains (like Wikipedia, Aspell, and Birkbeck datasets) shows that we are able to at least partially capture fundamental (and not dataset-specific) statistics about human spelling mistakes. In the next section we challenge this conclusion further, attempting to apply our method in non-English domains where no labeled data is available.

\subsection{Transfer to Resource-Scarce Languages}
Our procedure of training data generation is based on introduction of noise to natural language and relies solely on pre-computed typo-related statistics. Under a bold assumption that such statistics are largely language-agnostic, we show that it is possible to train a denoising transformer-based spelling correction model in settings where no \texttt{<typo\_string, correct\_string>} pairs are available. Leaving other statistics unchanged, we convert the character confusion matrix from English language to a target language using a QWERTY keyboard mapping. This way each English character is mapped to a character on the same keyboard key used in a target language layout. Using updated statistics, we train simple models for Russian\footnote{\url{https://github.com/Koziev/NLP_Datasets/blob/master/Samples/prep\%2Bnoun.zip}}, Arabic \cite{aly-atiya-2013-labr}, Greek \footnote{\url{https://repositori.upf.edu/handle/10230/19963}}, and Setswana \footnote{\url{https://repo.sadilar.org/handle/20.500.12185/404}} languages. Datasets for Arabic, Greek, and Setswana  are split into individual tokens. Datasets for Greek and Russian are lowercased. We use the same model configuration and training procedure as in section \cref{sec:real_vs_uni}. In Table \ref{tab:language_transfer} we report the number of unique examples and tokens for each dataset, alongside with sequence-level accuracy on a test set (not seen by BPE tokenizer and the model during training).

\begin{table}[H]
\centering
\begin{tabular}{lccc}
    \toprule
    Dataset&Example \#&Token \#&Accuracy\\
    \midrule
    Arabic&4,096,407&318,521&83.33\\
    Greek&9,491,753&270,269&93.97\\
    Russian&2,679,222&324,867&91.83\\
    Setswana&2,348,161&61,382&94.48\\
    \bottomrule
\end{tabular}
\caption[Language transfer results.]{Language transfer results.}
\label{tab:language_transfer}
\end{table}

Results indicate that this simple approach proves itself useful for bootstrapping spelling correction systems in settings when no labels are available. These findings may be especially helpful for languages suffering from scarcity of available resources, such as the majority of languages in Africa \cite{martinus2019focus}.

\section{Production Usage}
Trained model is loaded in memory using the TensorFlow \cite{abadi2016tensorflow} SavedModel format, and is fed all input strings shorter than \texttt{MAX\_INPUT\_LENGTH=20}. We limit max input size in order to ignore abnormally long inputs and to provide latency guarantees, as transformer time complexity is quadratic in the input sequence length.

\begin{figure}[htbp]
\resizebox{\columnwidth}{!}{%
\includegraphics{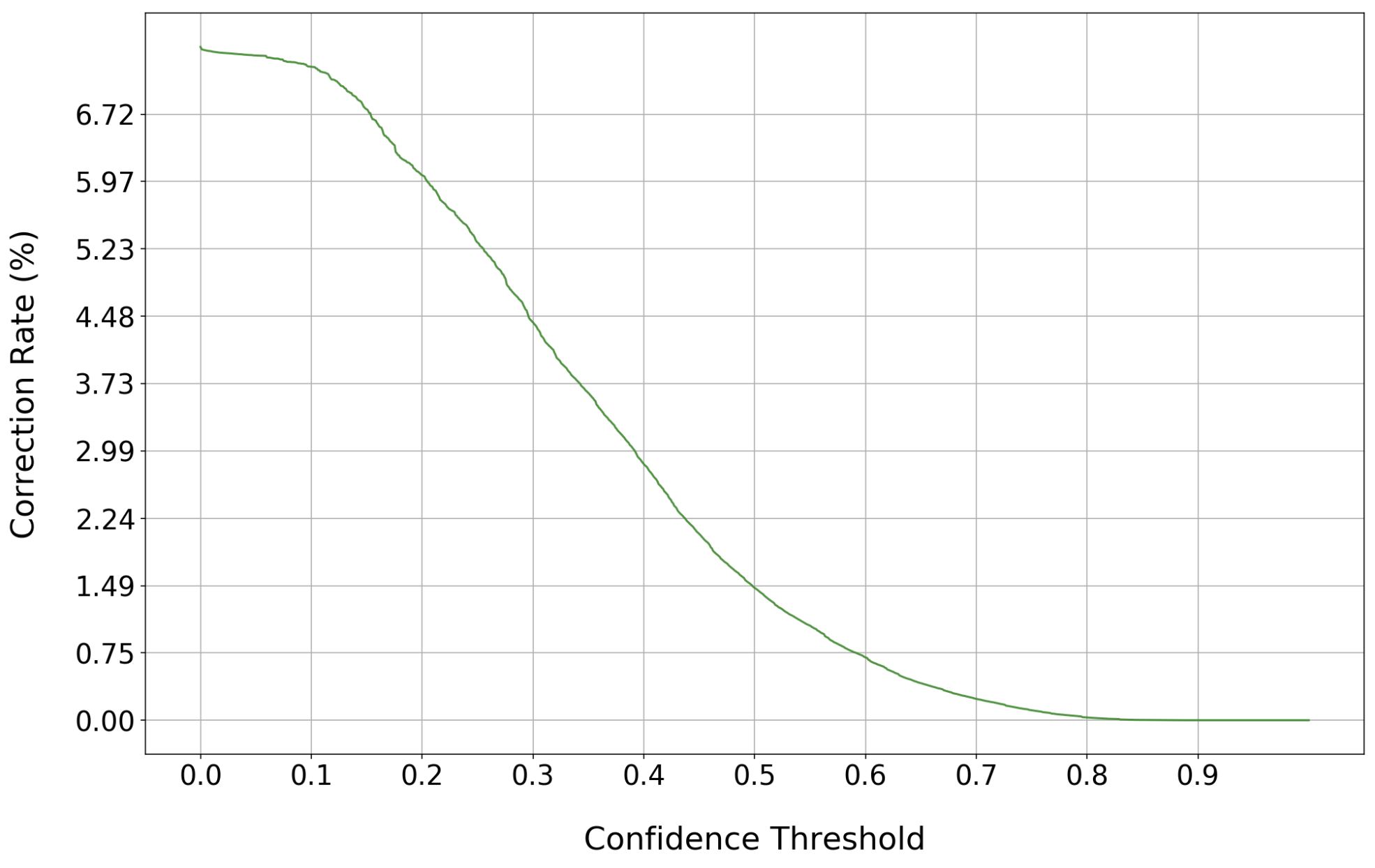}%
}
\caption{Correction Rate vs Confidence Threshold.}
\label{fig:correction_rate}
\end{figure}

Beam search of size 2 is performed when selecting top output sequence candidate, and we find that increasing beam size gives only minimal quality improvements at the expense of significantly higher latency. Beam search score is treated as a confidence score and is assigned to every prediction. Empirically chosen cut-off of 0.5 is used for serving predictions (i.e. all spelling corrections with score below this threshold are ignored), resulting in ~1.5\% of queries being corrected. Relationship between confidence threshold and spelling correction rate on {\ifaclfinal{HubSpot}\else{\textit{Company Name}}\fi} search logs is presented in Figure~\ref{fig:correction_rate}.

\section{Future Work}
This paper presents our first iteration on deep learning spelling correction, with multiple avenues for further improvement and research. In particular, we leave several items for future work:
\begin{itemize}
\item \textbf{Model architecture improvements.} Currently, we use a default transformer implementation, and there may be benefit in increasing model capacity, vocabulary and beam search size, using custom architectures, as well as a combination of models suggested by \newcite{li-etal-2018-hybrid}. Additional techniques like label smoothing, checkpoint averaging, and pretraining on a larger corpus may also improve model performance.
\item \textbf{Personalised recommendations.} Taking user context into account is key to providing a personalised search experience. Our current model is global and ignores user preferences. Embedding user context and using it as a feature may be an appropriate solution for this problem. Model architecture and findings from Gmail Smart Compose \cite{chen2019gmail} may be applicable here.
\item \textbf{Smarter noise generation.} Our current approach to typo generation is better than random but is still far from being perfect at emulating human behavior. For instance, \textit{Insertion} errors depend on both previous and next (relative to the injected character) characters. This is currently not taken into account. Additionally, we have very limited knowledge on how the probability of making a typo changes with the length of the string. Although known to be challenging, generative adversarial models for text \cite{fedus2018maskgan} may be used in order to generate errors indistinguishable from those of humans.
\end{itemize}

\section{Conclusion}
We presented a novel method for spelling correction - a denoising autoencoder transformer based on a noise generation procedure which generates artificial spelling mistakes in a realistic manner. Our contributions are three-fold, we: 1) demonstrated that a realistic typo generation procedure is superior to adding noise in a uniform way, 2) presented a way to train a spelling correction model in resource-scarce settings where no labeled data is available, and 3) by using unprocessed search logs showed that training a model directly on data from the target domain is possible and prevents the model from overcorrecting.

\bibliography{library}
\bibliographystyle{acl_natbib}

\end{document}